\documentclass[a4paper, 10pt, conference]{ieeeconf}      % Use this line for a4
\usepackage{FG2019}

\FGfinalcopy % *** Uncomment this line for the final submission

\usepackage{times}
\usepackage{epsfig}
\usepackage{graphicx}
\usepackage{amsmath}
\usepackage{amssymb}
\usepackage{caption}
\usepackage{subcaption}
\usepackage{footmisc}
\usepackage{pdfpages}

\IEEEoverridecommandlockouts                              % This command is only
\def\FGPaperID{205} % *** Enter the FG 2019 Paper ID here

\title{\LARGE \bf
Matching Thermal to Visible Face Images Using a Semantic-Guided Generative Adversarial Network
}

\author{\parbox{16cm}{\centering
    {\large Cunjian Chen and Arun Ross}\\
    {\normalsize
    Michigan State University, USA\\}}
}

\begin{document}
\IEEEoverridecommandlockouts\pubid{\makebox[\columnwidth]{978-1-7281-0089-0/19/\$31.00~\copyright{}2019 IEEE \hfill}
\hspace{\columnsep}\makebox[\columnwidth]{ }}

\ifFGfinal
\thispagestyle{empty}
\pagestyle{empty}
\else
\author{Anonymous FG 2019 submission\\ Paper ID \FGPaperID \\}
\pagestyle{plain}
\fi
\maketitle

%%%%%%%%%%%%%%%%%%%%%%%%%%%%%%%%%%%%%%%%%%%%%%%%%%%%%%%%%%%%%%%%%%%%%%%%%%%%%%%%
\begin{abstract}
Designing face recognition systems that are capable of matching face images obtained in the thermal spectrum with those obtained in the visible spectrum is a challenging problem. In this work, we propose the use of semantic-guided generative adversarial network (SG-GAN) to automatically synthesize visible face images from their thermal counterparts. Specifically, semantic labels, extracted by a face parsing network, are used to compute a semantic loss function to regularize the adversarial network during training. These semantic cues denote high-level facial component  information associated with each pixel. Further, an identity extraction network is leveraged to generate multi-scale features to compute an identity loss function. To achieve photo-realistic results, a perceptual loss function is introduced  during network training to ensure that the synthesized visible face is perceptually similar to the target visible face image. We extensively evaluate the benefits of individual loss functions, and combine them effectively to learn the mapping from thermal to visible face images. Experiments involving two multispectral face datasets show that the proposed method achieves promising results in both face synthesis and cross-spectral face matching. 
\end{abstract}

%%%%%%%%% BODY TEXT
\section{Introduction}
Matching thermal spectrum (THM) face images against visible spectrum (VIS) face images has received increased attention in the literature, due to its broad applications in the military, commercial, and law enforcement domains~\cite{HuSurvey2017}.  Thermal emissions from the face images are less sensitive to changes in ambient lighting. Further, thermal face images can be acquired in dark environments characterized by low ambient lighting thereby making them suitable for nighttime face recognition.  However, images present in legacy face datasets are  typically RGB images acquired in the visible spectrum. Therefore, cross-spectral matching of THM face images against VIS face images is of particular importance in delivering nighttime face recognition systems with a high degree of accuracy. 

The appearance variation between two face samples of the same subject captured under different illumination conditions can be larger than that of two samples belonging to two different subjects~\cite{CHEN201625}. Existing approaches for cross-spectral face matching can be categorized as follows: (a) photometric normalization of images in each spectral band~\cite{CHEN201625, Klare:2013:HFR}; (b) projecting THM and VIS images into a common subspace~\cite{Sarfraz2017, Seyed2018}; and (c) mapping THM images to VIS images via image synthesis~\cite{Riggan2016, Riggan2018_region}. These approaches have demonstrated effectiveness in minimizing inter-spectral differences and resulting in modest gains for cross-spectral face matching accuracy.  Their performance is still far from satisfactory for practical needs~\cite{HuSurvey2017} and the synthesized face images often appear unrealistic due to the lack of sufficient facial details~\cite{CHEN201625, Riggan2016}. 

Recent advances in generative adversarial networks (GANs) has witnessed success  in various face-related applications, including face completion~\cite{LiLY017}, frontalization~\cite{Tran2017, Huang_2017_ICCV, xiang_ffgan}, and age progression~\cite{Yang_2018_CVPR}. GANs are neural networks  that consist of two components: a generator that learns how to synthesize some type of data (e.g., images), and a discriminator that learns to discriminate  between real data and synthesized data~\cite{Goodfellow:2014:GAN}.  GANs have been used to learn the mapping from input face images to target face images, such as profile face images to frontal view face images~\cite{Huang_2017_ICCV}. The mapping function is typically constrained by the use of per-pixel loss functions computed between the output and target face images during training.  Loss functions are used during the training phase to measure the disparity between the actual output of the neural network and the expected target  output. GANs have also been used to synthesize VIS images from their THM counterparts~\cite{Zhang2017, Xing_2018_BTAS, Wang2018}. These solutions integrate additional loss functions, such as identity loss~\cite{Zhang2017}, attribute loss~\cite{Xing_2018_BTAS} and shape loss~\cite{Wang2018}, into the generative and discriminative components  in order to further constrain the mapping function. In spite of these advances, semantic information is not explicitly considered when using these GANs to learn the mapping from THM to VIS face images. We assert that these semantic cues, pertaining to the features around eyes, nose and mouth regions, may be beneficial for synthesizing identity-preserved face images.

In this regard, we propose a semantic-guided generative adversarial network (SG-GAN) to regularize GAN training with semantic priors in order to effectively synthesize VIS images from THM images. Specifically, the semantic priors are extracted by a face parsing network~\cite{LiuYHY15}. After that, these semantic priors are used to compute a semantic loss function. Similar schemes have been explored in the context of face completion~\cite{LiLY017} and deblurring~\cite{deblur2018}. However, the semantic priors have yet to be explored in the context of face recognition. Further, we show that extracting multi-scale identity features from a pre-trained face recognition network can boost the performance as well. 

\begin{figure*}[h]
  \centering
    \includegraphics[width=0.90\textwidth]{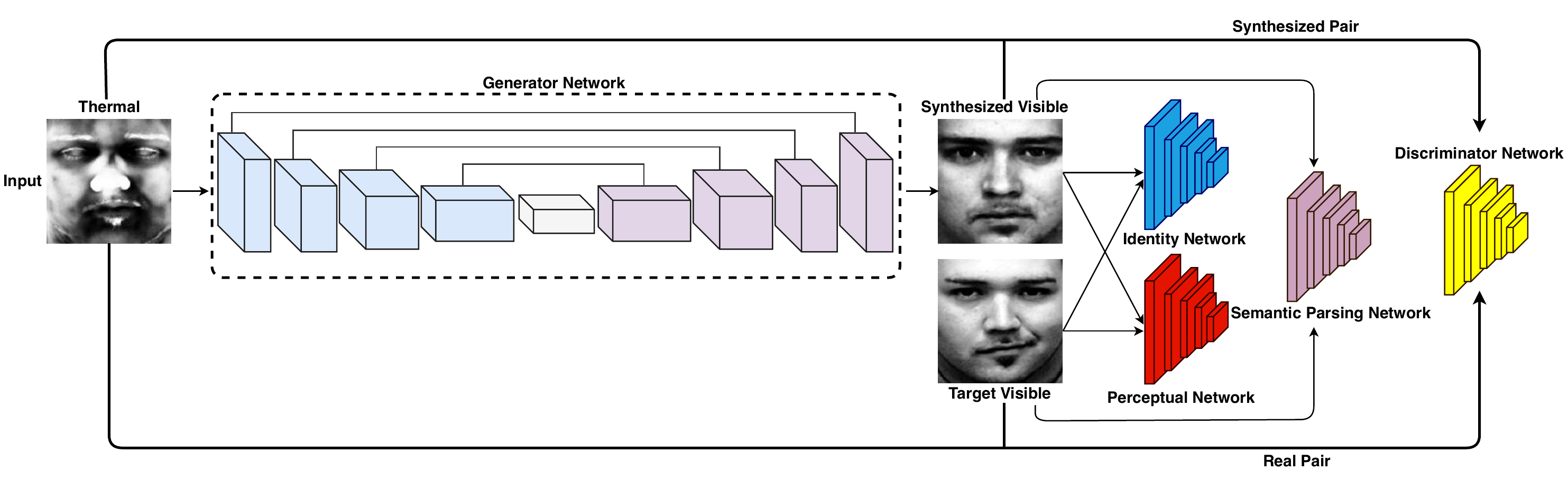}
    \caption{Flowchart depicting the training of the proposed SG-GAN framework. The generator takes a thermal image as an input to produce a synthesized visible image as an output. The discriminator is trained to distinguish between two pairs of images: the real pair, consisting of an input thermal image and a target visible image, and the synthesized pair, consisting of an input thermal image and a synthesized visible image. The identity and perceptual networks accept a synthesized visible image and the target visible image to compute the identity loss and perceptual loss values, respectively. The semantic parsing network accepts both synthesized visible image and the target visible image to compute the semantic loss value. }
    \label{sg-gan-network}
\end{figure*}

The framework of the proposed SG-GAN method is illustrated in Figure~\ref{sg-gan-network}. The generator network is an encoder-decoder architecture with skip-connections, whose input is a thermal image, and the output is a synthesized visible image. The discriminator network is a CNN classifier that learns to separate the real pairs from the synthesized pairs. The real pair consists of input thermal and target visible images  (ground-truth), and the synthesized pair consists of input thermal and synthesized visible images. The identity, perceptual and semantic parsing networks accept a synthesized visible image and the target visible image to compute the identity loss, perceptual loss and semantic loss, respectively.  A weighted combination of different loss functions is used to optimize the entire network during training. Only the generator network is required during the testing. 

The main contributions of this work are summarized here:
\begin{itemize}
\item We use a face parsing network to extract semantic labels as priors to regularize the mapping from thermal to visible face images.
\item We use an identity network to extract multi-scale identity features. 
\item We investigate different loss functions and demonstrate their individual as well as combined effectiveness. 
\item We evaluate the proposed network on two different benchmark multispectral face datasets and achieve promising results. 
\end{itemize}

The rest of the paper is organized as follows. Section~\ref{related_work} discusses recent advancements in synthesizing VIS face images from THM face images. Section~\ref{proposed_method} describes the proposed SG-GAN method used in this work, with a particular emphasis on the various loss functions. Section~\ref{exp_results} discusses the image synthesis and face matching results on two multispectral face datasets. Conclusions are drawn in Section~\ref{conclusion}. 

\section{Related Work}\label{related_work}
In this section, we briefly discuss the existing literature on synthesizing visible face images from thermal face images. 

\textbf{Feature-based Synthesis:} Chen and Ross~\cite{CHEN201625} demonstrated that a VIS image could be reconstructed from a THM image  using hidden factor analysis. Riggan et al.~\cite{Riggan2016} performed  VIS image reconstruction using a two-stage process, consisting of feature extraction and feature regression  using CNNs. However, these reconstructed results were observed to be blurry. In a subsequent work, Riggan et al.~\cite{Riggan2018_region}  used a fully convolutional neural network to learn a global mapping  between THM and VIS images, as well as the regions around the eyes, nose and mouth. The final synthesized image was a combination of global and local mapping functions resulting in better quality output. 

\textbf{GAN-based Synthesis:} With the advent of generative adversarial networks, image-to-image synthesis has achieved promising results. Zhang et al.~\cite{Zhang2017} proposed a GAN-based method to synthesize VIS face images from THM images. A combination of identity loss and perceptual loss functions was used to optimize the proposed framework. The identity loss function was computed by the features extracted from a single layer of a fine-tuned VGG model~\cite{VGG15}. Similarly, a closed-set face recognition loss function was proposed to regularize the discriminator network during training in~\cite{Zhang2018}. Their discriminator network not only distinguished  between real and synthesized  samples, but also performed closed-set face recognition. The face recognition loss function was computed using a pre-trained VGG model without any fine-tuning. In addition to the identity loss  function, the attribute loss function has also been used to optimize the network~\cite{Xing_2018_BTAS}. In~\cite{Xing_2018_BTAS}, an attribute predictor was developed by fine-tuning the VGG-Face network using 10 annotated attributes. The experiments demonstrated that the incorporation of the attribute loss function resulted in  much better performance than when using the identity loss function alone. Wang et al.~\cite{Wang2018} incorporated the shape loss function into the CycleGAN consisting of a generative network and a detector network. The shape loss value was computed as the euclidean distance of the 68 detected landmarks between the synthesized VIS image and the target VIS image. 

However, all the aforementioned methods do not explicitly considered the semantic information  of the face to regularize network training. The main difference between our proposed method and~\cite{Xing_2018_BTAS} is that we use a semantic loss function to regularize the training process, whereas the latter uses an attribute loss function to guide the learning process. Further, we demonstrate that the use of semantic loss function is able to reduce the per-pixel loss value, calculated between the synthesized VIS face and target VIS face. Another notable difference is that SG-GAN extracts multi-scale features using an identity feature extraction network (see Figure~\ref{sg-gan-network}). 

%------------------------------------------------------------------------
\section{Proposed Method}\label{proposed_method}
In this work, the proposed SG-GAN is developed based on an existing generative adversarial network model, known as pix2pix~\cite{pix2pix}. Given a thermal image $x$ of size $3\times H \times W$,  the objective of our method is to learn the mapping from $x$ to $y$ in a supervised setting, i.e., $G: x\rightarrow y$, where $y$ is the target visible image of size $3\times H \times W$. For the generator, we used U-Net~\cite{pix2pix} with skip connections, which concatenates all the channels at layer $i$ with those at layer $n$-$i$.  Here, $n$ is the total number of layers and $i$ is the specific layer number. This allows the low-level information to be shared between the encoder and the decoder layers (see Figure~\ref{sg-gan-network}). For the discriminator, $D$, we used the PatchGAN~\cite{pix2pix} classifier that operates on $70\times70$ image patches and classifies them as being real or synthesized. Both generator and discriminator use a sequence of convolution blocks characterized by a composite function of three different operations: convolution, Batch Normalization (BN), followed by a Rectified Linear Unit (ReLU). A dropout rate of 50\% is used for $G$. To optimize these two networks, the gradient update is alternated  after every step between $D$ and $G$. 

\subsection{Identity Extraction Network}
To extract identity features from face images, we train a face recognition network based on the VGG-19 network~\cite{VGG15} from scratch. Specifically, we utilize a newly created large-scale face dataset termed VGGFace2~\cite{vggface2} to train this network. MTCNN~\cite{ZhangZLQ16} is utilized to automatically detect the face and its landmark. The detected landmarks are used to geometrically normalize the face image to a size of $112\times96$. After that, the input image is cropped based on a randomly generated aspect ratio and resized to $224\times224$. The VGG-19 architecture has 16 convolutional layers and 3 fully-connected layers. In this setting, we do not use batch normalization to train the identity network. Training images were randomly flipped horizontally to facilitate data augmentation.  A starting learning rate of 0.001 and a batch size of 32 were used. The maximum number  of  epochs for training was set to  90. 

To better extract the identity-specific features, intermediate features from multiple layers of VGG-19 are concatenated together. Specifically, we have considered features extracted from the $relu1$-$1$, $relu2$-$1$, $relu3$-$1$, $relu4$-$1$ and $relu5$-$1$ layers. These concatenated features are used to compute the identity loss function between the synthesized VIS and target VIS face images. In the end, the final identity loss function is weighted by different coefficients, where larger weights are assigned to the features extracted at deeper layers of the network. The default weights for the individual layers are 1/32, 1/16, 1/8, 1/4, and 1, respectively. A similar approach was used to compute the perceptual loss function for high-resolution image synthesis in~\cite{Wang2017}. 

\subsection{Semantic Face Parsing Network}
\begin{figure}
  \centering
    \includegraphics[width=0.45\textwidth]{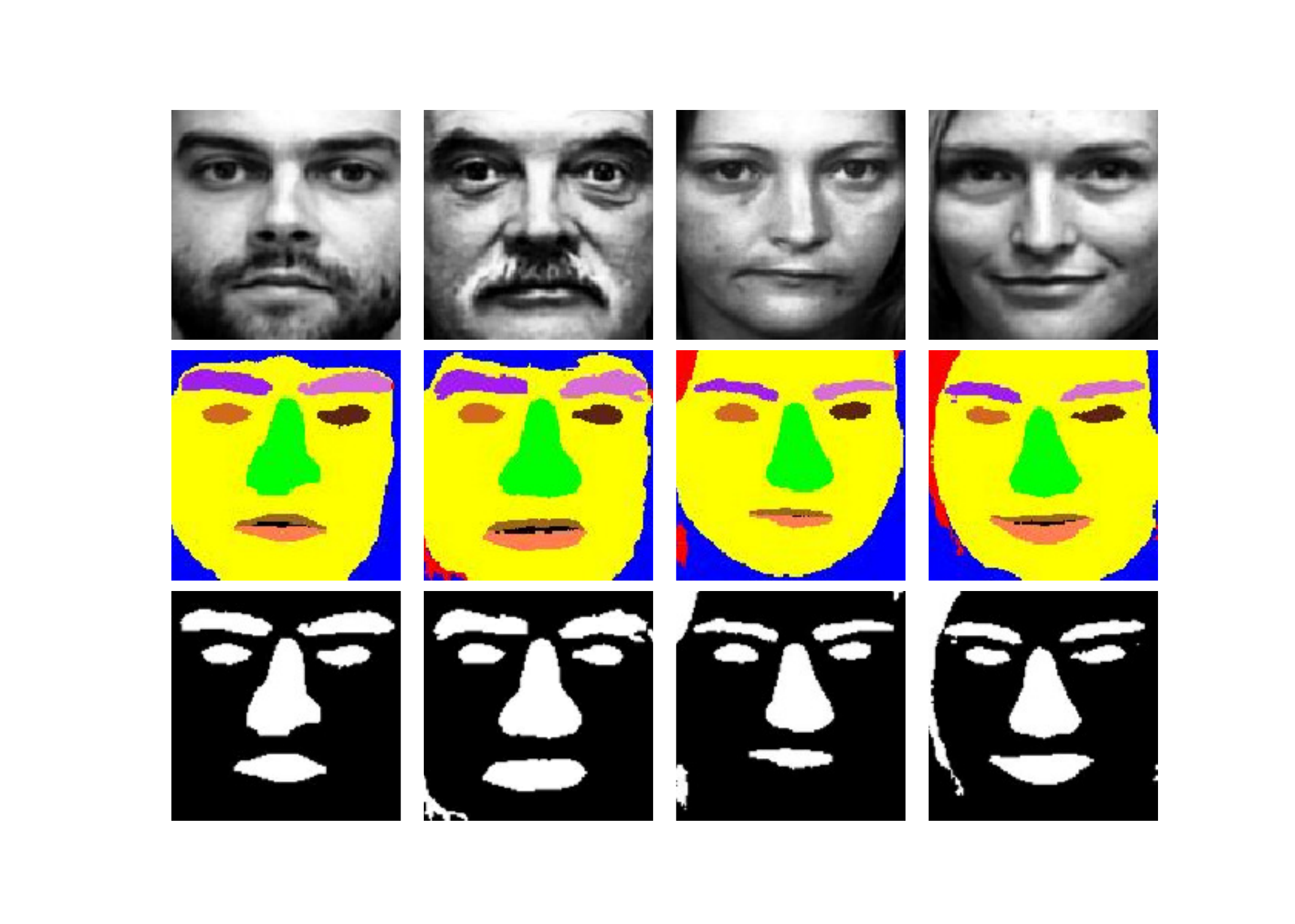}
    \caption{Examples of face parsing results. The images in the top row are the visible face images, while the ones in the middle are the 11-class semantic label images. The bottom
row is the result of converting the 11-class semantic label images to
2-class semantic label images. The visible face images are from the PCSO dataset~\cite{CHEN201625}. }
    \label{pcso_parsing}
\end{figure}

Existing literature on synthesizing VIS face images from THM face images does not explicitly consider the semantic information related to different facial components. In this approach, we propose to regularize the adversarial network training by introducing semantic priors as an additional loss function.\footnote{The thermal image has been converted from greyscale to RGB so that it conforms to the generator network's input.} Here, the semantic priors are calculated by a face parsing network~\cite{LiuYHY15}, where the input is the visible face image and the output is the semantic labels that correspond to 11 different classes (see Figure~\ref{pcso_parsing}). These classes, corresponding to different facial components, are based on the label information provided by the HELEN dataset~\cite{LiuYHY15}. These labeled facial components correspond to face skin, left eye, right eye, left brow, right brow, nose, inner mouth, upper lip, lower lip, hair, and background~\cite{SmithZBLY13}. Since we are interested in salient facial components that are crucial to face recognition, class labels associated with the eyebrows, eyes, nose and mouth regions are grouped together. The resulting semantic label image is used to compute the semantic loss function. Note that the thermal and visible face images are aligned based on the manually annotated eye landmarks. The semantic labels are calculated for both synthesized visible image and target visible image. Figure~\ref{pcso_parsing} shows the semantic parsing results on visible face images. It must be noted that these parsing results can be further refined to remove outliers (e.g., part of the background is incorrectly classified as face skin in Figure~\ref{pcso_parsing}), by fine-tuning the pre-trained model on the benchmark datasets used in this work. 

\subsection{Loss Functions}
SG-GAN uses a set of loss functions that consists of adversarial and per-pixel loss values, as well as identity, perceptual and semantic loss values. These loss functions are independently investigated and then combined in an effective manner. 
\subsubsection{Adversarial Loss}
The adversarial loss function, $\mathcal{L}_{GAN}$, is defined as~\cite{pix2pix}:
\begin{equation}
\begin{split}
\mathcal{L}_{GAN}(G,D) & = \mathbb{E}_{x,y \sim p_{data}(x,y)}[\log D(x,y)]  \\
                                        & +  \mathbb{E}_{x \sim p_{data}(x)}[\log (1-D(x, G(x)))],\\
\end{split}
\end{equation}
where $G$ is the generator, $D$ is the discriminator, $x$ is the thermal image and $y$ is the target visible image. $p_{data}(x)$ indicates that $x$ is from the true data distribution and $p_{data}(x,y)$ indicates that both $(x,y)$ are from the true data distribution. The objective of generator $G$ is to synthesize visually realistic visible face images from thermal face images, while the discriminator $D$ is structured to distinguish the target visible face images from the synthesized ones, conditioned on the input thermal image. This min-max game will reach an equilibrium when neither $G$ nor $D$ can further reduce their loss values~\cite{gan_landscape}. In summary, $G$ attempts to minimize the objective function while $D$ attempts to maximize it. Mathematically, this could be described as: 
\begin{equation}
G^{*}= \arg \min_{G}\max_{D}\mathcal{L}_{GAN}(G,D).
\end{equation}
In practice, the generator maximizes the probability of synthesized VIS samples to be classified as real  by the discriminator. Therefore, the above loss function can be alternatively updated as~\cite{gan_landscape}:
\begin{equation}\label{GAN_D}
\begin{split}
\mathcal{L}_{D} & =  -\mathbb{E}_{x,y \sim p_{data}(x,y)}[log D(x,y)]\\
& - \mathbb{E}_{x \sim p_{data}(x)} [log(1-D(x, G(x)))],
\end{split}
\end{equation}
and
\begin{equation}\label{GAN_G}
\mathcal{L}_{G}= \mathbb{E}_{x \sim p_{data}(x)} [-log(D(x,G(x)))].
\end{equation}
Here, $\mathcal{L}_{G}$ is the generator loss and $\mathcal{L}_{D}$ is the discriminator loss. Maximizing $D$ is the same as maximizing $log(D)$. An optimal convergence of the adversarial network will result in $\mathcal{L}_{D}(real)=\mathcal{L}_{D}(synthesized)=0.5$ and $\mathcal{L}_{G}=0$~\cite{Luo_2018_ECCV}. This would indicate that the discriminator is unable to separate the synthesized pairs from the real pairs. 

\subsubsection{Per-pixel Loss  Function}\label{l1_loss}
The per-pixel loss function, $\mathcal{L}_R$, is computed as the mean value of the absolute element-wise difference between the images:
\begin{equation} \label{eu_eqn}
\mathcal{L}_R(G)=\mathbb{E}_{x,y \sim p_{data}(x,y)} || G(x)-y ||_{1},
\end{equation}
where, $G(x)$ and $y$ are the synthesized and target visible images, respectively.  $\mathcal{L}_R$ is used to reduce the space of all possible mapping functions between THM and VIS images such that the synthesized VIS samples  look visually similar to the target VIS images. We also tried replacing $\mathcal{L}_R$ with smooth $L_1$ or $L_2$ norms, but observed no significant improvement in matching performance. The per-pixel loss function compares the synthesized and target visible images at pixel-level. 

\subsubsection{Perceptual Loss Function}
The perceptual loss function, $\mathcal{L}_P$, is defined as follows,
\begin{equation} \label{eu_eqn}
\mathcal{L}_P(G)=\mathbb{E}_{x,y \sim p_{data}(x,y)} || \phi_{P} (G(x))-\phi_{P} (y)||_{1},
\end{equation}
where, $\phi_{P}$ denotes the features extracted from multiple layers of the pre-trained VGG-19 network on ImageNet dataset. The perceptual loss function~\cite{Justin2016} is used to measure the high-level semantic difference between the synthesized visible face image and the target visible face image; this ensures that the synthesized result is smooth and perceptually similar  to the target. The features are extracted at multiple layers and concatenated together to form a single feature descriptor. 

\subsubsection{Identity Loss Function}
The identity loss function, $\mathcal{L}_I$, is defined as follows:
\begin{equation} \label{eu_eqn}
\mathcal{L}_I(G)=\mathbb{E}_{x,y \sim p_{data}(x,y)} || \phi_{I}(G(x))-\phi_{I}(y)||_{1},
\end{equation}
where, $\phi_{I}$ denotes the features extracted from multiple layers of the pre-trained VGG-19 network on face dataset.  The $\mathcal{L}_I$ loss function is used to ensure that the synthesized visible face image contains identity-specific  features that are similar to the ground-truth target visible image. Although the use of $\mathcal{L}_R$ loss function, as defined earlier, can ensure visual similarity between the synthesized visible image and the real original image, it could produce blurry results since $\mathcal{L}_R$ tends to smoothen the output image. $\mathcal{L}_R$ also lacks high-level information about the identity. Hence, it is of particular importance to integrate the identity loss function $\mathcal{L}_I$ to extract high-level identity-specific features. 

\subsubsection{Semantic Loss Function}
The semantic loss function, $\mathcal{L}_S$, is defined as follows:
\begin{equation} \label{eu_eqn}
\mathcal{L}_S(G)=\mathbb{E}_{x,y \sim p_{data}(x,y)} || \phi_{S}(G(x))-\phi_{S}(y)||_{1},
\end{equation}
where, $\phi_{S}$ denotes the semantic labels extracted from the pre-trained semantic parsing network. The $\mathcal{L}_S$ loss function is used to ensure that the  shape and size of the synthesized visible face image is consistent with that of the ground-truth target visible image. The semantic loss value is measured as the difference of the semantic labels computed between the synthesized visible and target visible images. 

\subsection{Implementation Details}
The loss function used in our proposed SG-GAN framework was formulated as follows: 
\begin{equation} \label{loss_eqn}
\begin{split}
\mathcal{L} & = \lambda_G\mathcal{L}_{GAN}(G,D)+\lambda_R\mathcal{L}_R(G) \\
                  & + \lambda_P \mathcal{L}_P(G)+\lambda_I \mathcal{L}_I(G)+\lambda_S \mathcal{L}_S(G),
\end{split}
\end{equation}
where, $\mathcal{L}_{GAN}$ is the adversarial loss consisting of both generator and discriminator losses, $\mathcal{L}_R$ is the per-pixel loss between synthesized visible face image and the target visible face image, $\mathcal{L}_P$ is the perceptual loss,  $\mathcal{L}_I$ is the identity loss and $\mathcal{L}_S$ is the semantic loss. Based on empirical analysis, we set $\lambda_G=1$, $\lambda_R=100$, $\lambda_P=10$, $\lambda_I=20$ and  $\lambda_S=1$ as the default weights for $\mathcal{L}_{GAN}$, $\mathcal{L}_R$, $\mathcal{L}_P$, $\mathcal{L}_I$ and $\mathcal{L}_S$, respectively. The final objective function is:
\begin{equation}
\begin{split}
G^{*} & = \arg \min_{G}\max_{D} \lambda_G\mathcal{L}_{GAN}(G,D)+\lambda_R\mathcal{L}_R(G)\\
         & + \lambda_P \mathcal{L}_P(G)+\lambda_I \mathcal{L}_I(G)+\lambda_S \mathcal{L}_S(G).
\end{split}
\end{equation}
The proposed framework was  implemented in PyTorch. During the training phase, batch normalization with Adam optimization was used. The default number of epochs used in our training was 200. Random cropping and image flipping operations  were used for  data augmentation  during training. We first scale the images to size $286\times286$, and then randomly crop them to a size of $256\times256$. The starting learning rate for Adam optimization was  0.0002, which was  fixed for the first 100 epochs and decreased by 1/100 after each subsequent epoch. The remaining parameters use the default values described in~\cite{pix2pix}.   
%We do not apply LSGANs~\cite{pix2pix} for training; hence, the binary cross entropy loss function is used to compute the difference between the target and the output for the adversarial loss. 

%------------------------------------------------------------------------
\section{Experimental Results}\label{exp_results}
To verify the effectiveness of the proposed method, experiments were conducted on the Pinellas County Sheriff's Office (PCSO)~\cite{CHEN201625} and Army Research Laboratory (ARL)~\cite{HuDatabase2016} datasets. It must be  noted that the PCSO and ARL datasets consist of face images acquired in the middle-wave infrared (MWIR) and long-wave infrared (LWIR) spectra, respectively. To evaluate face matching performance, we choose face recognition matchers that have achieved state-of-the-art results on the LFW dataset~\cite{AMSoftmax2018, ChenLGH18}. 

\subsection{Face Recognition Matchers}
\textbf{AM-Softmax:} The AM-Softmax~\cite{AMSoftmax2018} matcher was trained on the VGGFace2~\cite{vggface2} dataset consisting of 7,773 subjects and 1,428,908 images.  AM-Softmax was developed by introducing additive angular margin for the Softmax loss after performing both feature and weight normalization. We used stochastic gradient descent (SGD) with momentum to update the weights.  The momentum was set  to 0.9 and the base learning rate was initialized to 0.1. The batch size was 256 and the maximum number of iterations was 30,000. We used the ``step'' learning rate policy which drops the learning rate by a factor of 0.1 after 16,000, 24,000 and 28,000 iterations. This results in a trained model of size 106MB. After that, we extracted a 1024-dimensional feature vector from the penultimate fully connected layer to represent the input  image. The match score was computed using the cosine similarity metric. We evaluated the trained model on the LFW benchmark dataset and achieved a  classification accuracy of 99.35\%, thereby suggesting the efficacy of this method for visible spectrum face recognition. 

\textbf{MobileFaceNet:} MobileFaceNet~\cite{ChenLGH18}, a reminiscent of MobileNetV2~\cite{Sandler_2018_CVPR}, uses global depth-wise convolution layer to replace the global average pooling layer. It significantly reduces the model size while maintaining comparable recognition accuracies on the LFW and MegaFace datasets~\cite{ChenLGH18}. We employed the MobileFaceNet architecture and trained using the angular softmax (A-Softmax) loss function~\cite{sphereface2017} on the VGGFace2 dataset. The momentum was set  to 0.9 and the base learning rate was initialized to 0.1. The batch size was 128 and the maximum number of iterations was set to 60,000. We used the ``step'' learning rate policy which drops the learning rate by a factor of 0.1 after 36,000, 50,000 and 58,000 iterations. This resulted in a trained model size of 8MB. After that, we extracted a 256-dimensional feature vector from the penultimate fully connected layer and used the cosine similarity metric  to compare feature vectors. We tested the model on the LFW benchmark dataset and achieved a classification accuracy of 99.30\%. 

The AM-Softmax and MobileFaceNet matchers deliver strong baselines. Considering that the VGGFace network is employed in this work to extract identity-specific features during SG-GAN training, we use these other matchers to demonstrate the generalization capability of the SG-GAN in synthesizing VIS from THM face images. In this setting, the synthesized visible image is matched against the visible image. 

%-------------------------------------------------------------------------
\subsection{Evaluation on PCSO Dataset}
\begin{figure*}[h]
  \centering
    \includegraphics[width=0.5\textwidth]{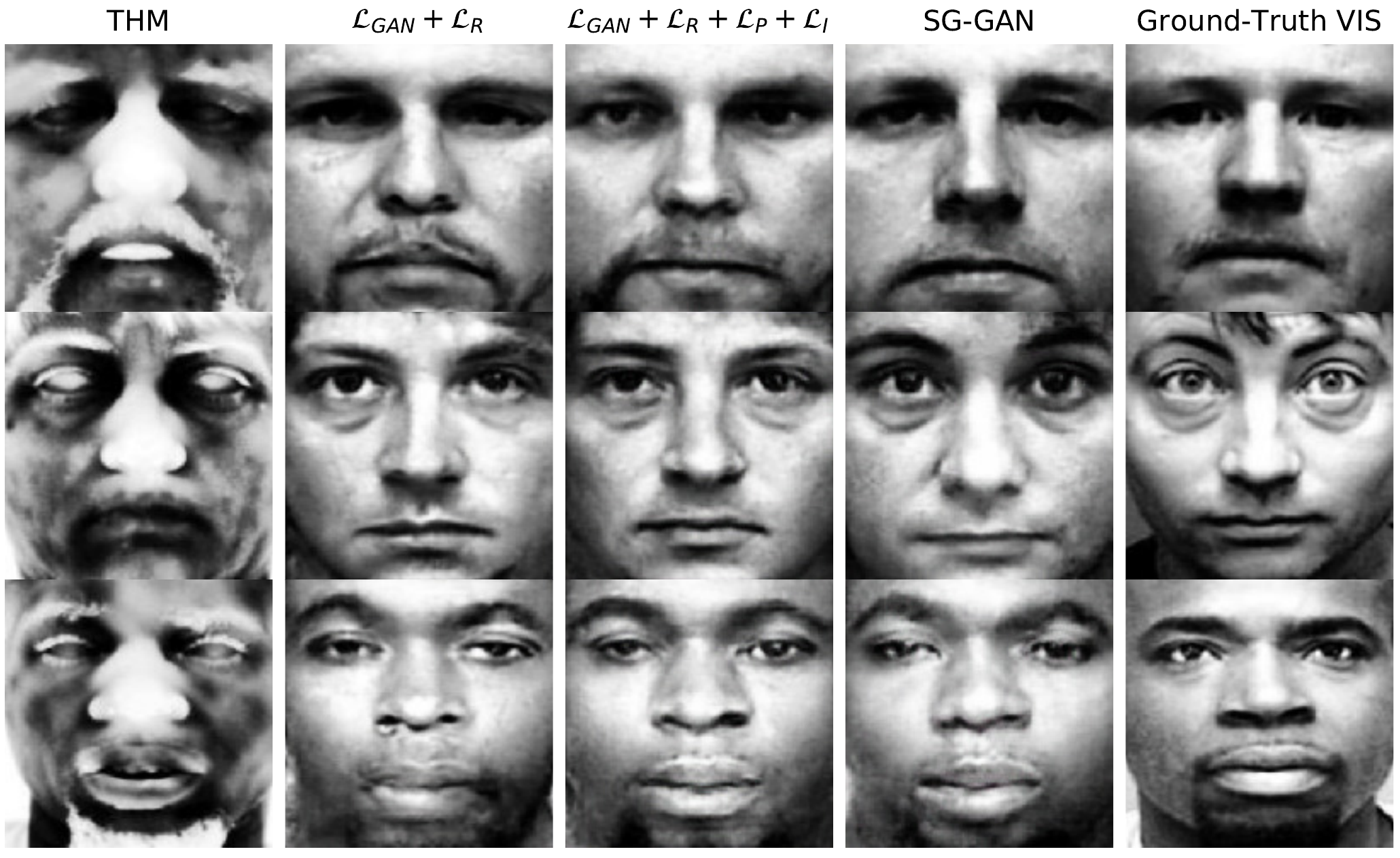}
    \caption{Synthesizing VIS face images from THM images on the PCSO dataset.  Compared to the use of the ``$\mathcal{L}_{GAN}+\mathcal{L}_R+\mathcal{L}_P+\mathcal{L}_I$'' loss function, the output of SG-GAN is semantically more close to the ground-truth VIS image especially around the salient facial regions.}
    \label{pcso_synthesized}
\end{figure*}

The PCSO dataset contains data from 1,004 subjects, where 1,003 of the subjects have two visible face images and one thermal face image each. Based on manually localized eye coordinates, the face images were aligned and cropped to an image size of $150\times130$. Following the evaluation benchmark given in~\cite{CHEN201625}, the first 667 subjects were used for training and the rest were used for testing. The training and test subsets consist of 1,333 and 337 THM-VIS pairs, respectively. There is no overlap between the training and test subjects. Examples of thermal, synthesized visible and ground-truth visible face images are shown in Figure~\ref{pcso_synthesized}. The proposed SG-GAN method appears to have produced photo-realistic results. Discriminative features pertaining to the eye, nose and mouth regions are better preserved. This suggests that it is beneficial to embed semantic priors when training the GAN. In this regard, we have considered the semantic class labels related to eyebrows, eyes, nose and mouth regions. Further, the use of the identity extraction network allows identity-specific features to be similar to that of the targets. The corresponding face matching experiments, reported with Area Under the Curve (AUC) and Equal Error Rate (EER), are summarized in Table~\ref{PCSO_results}.

\begin{table}[h]
\scriptsize
\caption{Evaluation of face matching using  the  AM-Softmax and MobileFaceNet matchers on the PCSO dataset. The impact of different loss functions can be seen here. A higher AUC is better while a lower EER is better. }
\begin{center}
  \begin{tabular}{ | c | c | c |  c | c | c |}
    \hline
    & \multicolumn{2}{c|}{\textbf{AM-Softmax}} & \multicolumn{2}{c|}{\textbf{MobileFaceNet}}   \\ \hline
     & \textbf{AUC (\%)} & \textbf{EER (\%)} & \textbf{AUC (\%)} & \textbf{EER (\%)} \\ \hline
    Direct Matching & 69.98 & 34.65  & 69.18 & 35.52 \\ \hline
    $\mathcal{L}_{GAN}$+$\mathcal{L}_R$~ & 86.73 & 21.36 & 87.59 & 20.77\\ \hline
     $\mathcal{L}_{GAN}$+$\mathcal{L}_R$+$\mathcal{L}_P$ & 87.53 &19.82 & 88.47 & 19.09\\ \hline
    $\mathcal{L}_{GAN}$+$\mathcal{L}_R$+$\mathcal{L}_P$+$\mathcal{L}_I$ & 89.76 &18.40 & 91.14 &17.51\\ \hline
    Proposed (SG-GAN) & \textbf{90.12} & \textbf{15.98} & \textbf{92.16} & \textbf{15.01} \\ 
    \hline
  \end{tabular}
\end{center}
\label{PCSO_results}
\end{table}

\begin{figure*}[h]
    \centering
    ~ %add desired spacing between images, e. g. ~, \quad, \qquad, \hfill etc. 
    \begin{subfigure}[b]{0.40\textwidth}
        \includegraphics[width=\textwidth]{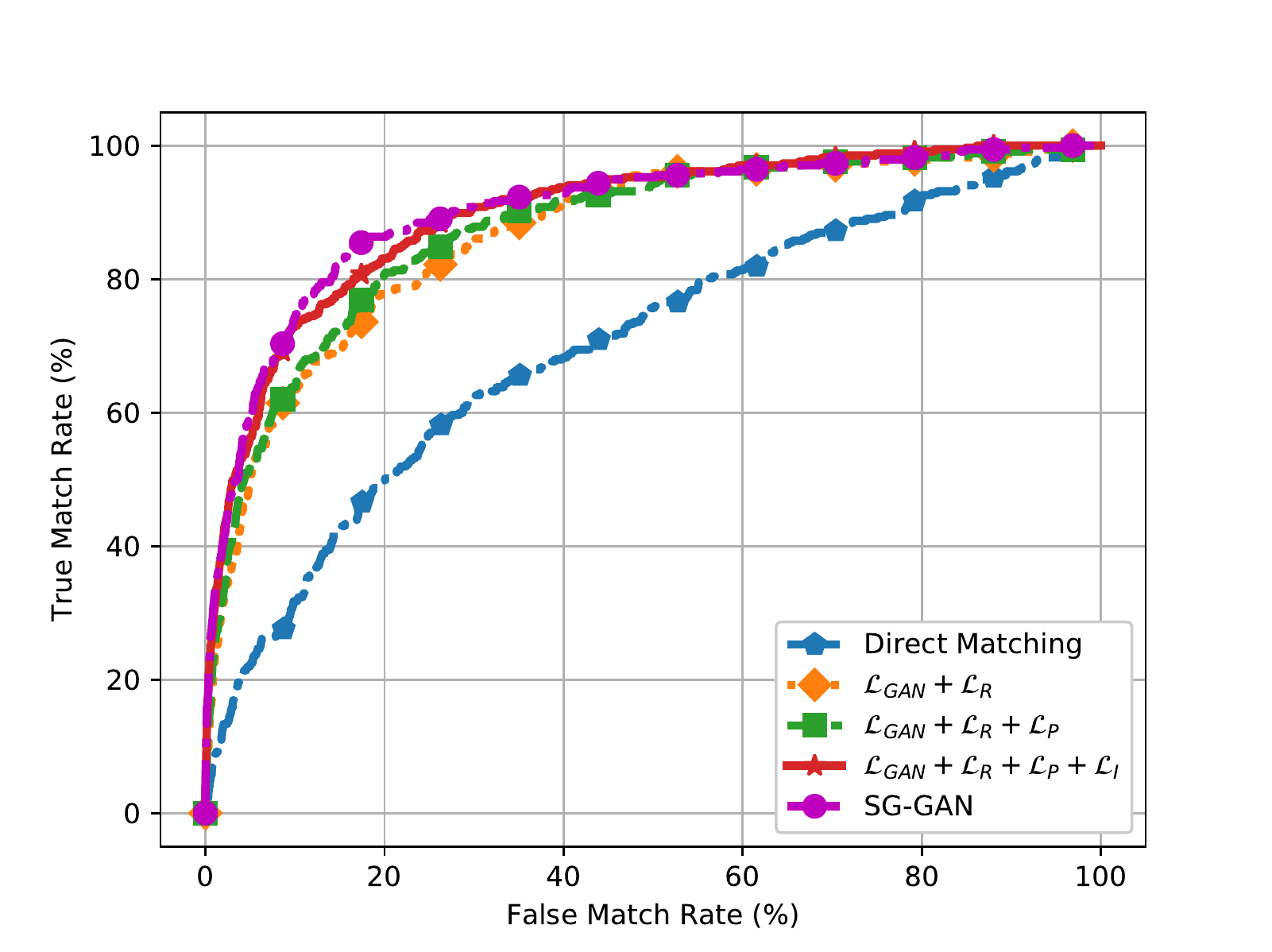}
        \caption{AM-Softmax}
        \label{pcso_roc}
    \end{subfigure}
    ~ %add desired spacing between images, e. g. ~, \quad, \qquad, \hfill etc. 
    \begin{subfigure}[b]{0.40\textwidth}
        \includegraphics[width=\textwidth]{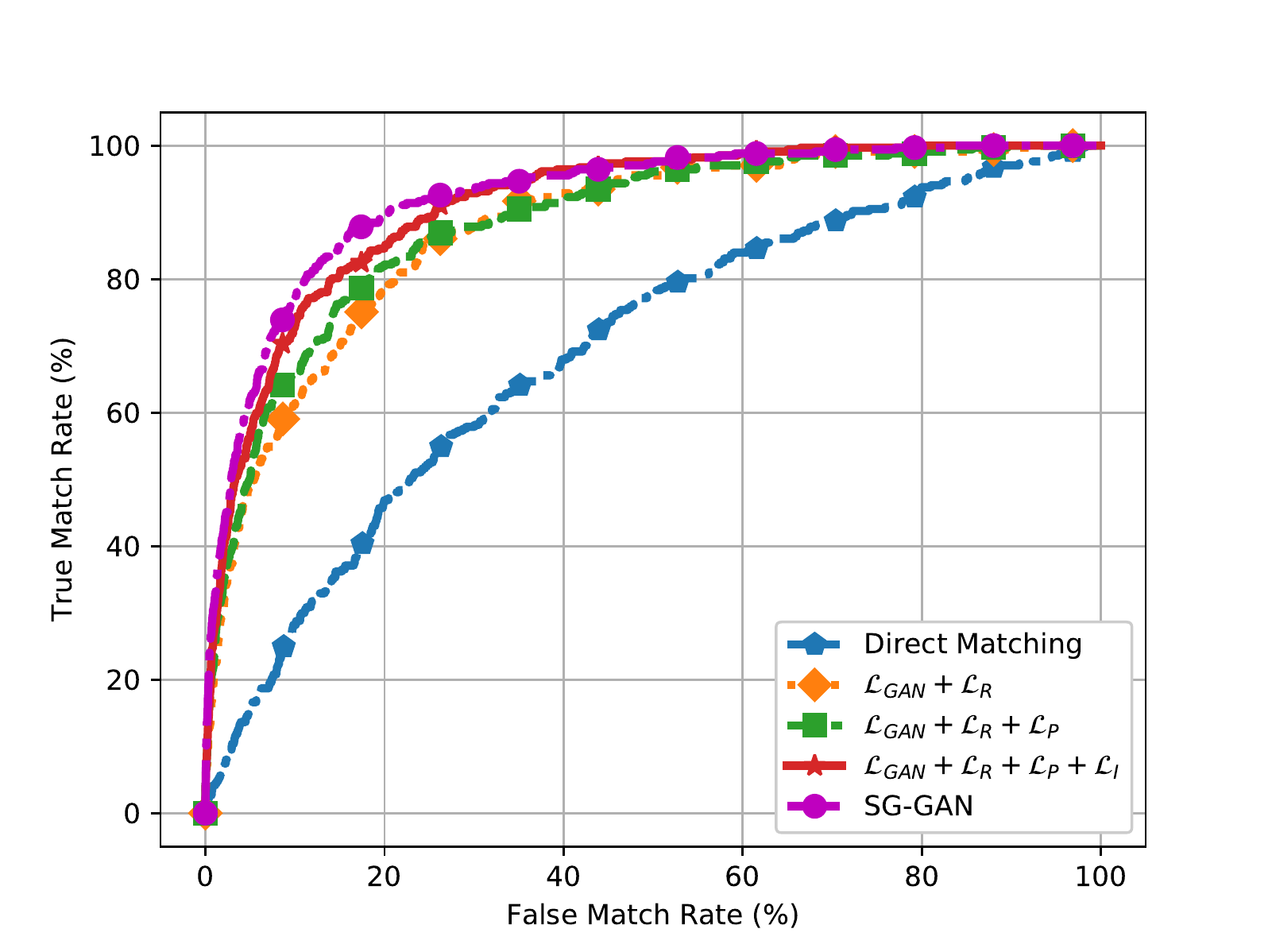}
        \caption{MobileFaceNet}
        \label{pcso_roc_mobilefacenet}
    \end{subfigure}
    \caption{An ablation study of different loss functions on the PCSO dataset using different face matchers. Our proposed solution,  SG-GAN, achieves much better performance than $\mathcal{L}_{GAN}+\mathcal{L}_R$. The figure is best viewed in color.}\label{pcso_roc_matchers}
\end{figure*}

{\bf Ablation Study:} To demonstrate the effectiveness of different loss functions, an ablation study was conducted using the PCSO dataset. As can be seen in Table~\ref{PCSO_results} and Figure~\ref{pcso_roc_matchers}, the perceptual loss function $\mathcal{L}_P$ contributes slightly to the improvement of face matching performance. The addition of $\mathcal{L}_I$ loss function results in a pronounced difference in the AUC value. Here, ``Direct Matching'' refers to the setting where (a) deep features are directly extracted from THM and VIS face images and (b) the extracted features are compared to produce a match score. ``$\mathcal{L}_{GAN}+\mathcal{L}_R$'' denotes the performance of the original pix2pix model~\cite{pix2pix}. $\mathcal{L}_{GAN}$+$\mathcal{L}_R$+$\mathcal{L}_P$ refers to the performance where perceptual loss function $\mathcal{L}_P$ is added. Similarly, $\mathcal{L}_{GAN}$+$\mathcal{L}_R$+$\mathcal{L}_P$+$\mathcal{L}_I$ is used to denote the matching performance when both $\mathcal{L}_P$ and $\mathcal{L}_I$ loss functions are added. The proposed SG-GAN method is developed on the basis of $\mathcal{L}_{GAN}$+$\mathcal{L}_R$+$\mathcal{L}_P$+$\mathcal{L}_I$ that has been further regularized by a semantic loss function that is computed from the semantic labels extracted via the face parsing network. 

%\begin{figure}
%  \centering
%    \includegraphics[width=0.45\textwidth]{Figures/visualization_loss_pcso.eps}
%    \caption{Visualization of different loss functions with and without the semantic regularization on PCSO dataset. The top row shows the $\mathcal{L}_1$ loss and the bottom row shows the $\mathcal{L}_I$ loss.}
%    \label{visualization_loss_pcso}
%\end{figure}

\begin{figure*}[h]
    \centering
    \begin{subfigure}[t]{0.45\textwidth}
        \centering
        \includegraphics[width=0.9\textwidth]{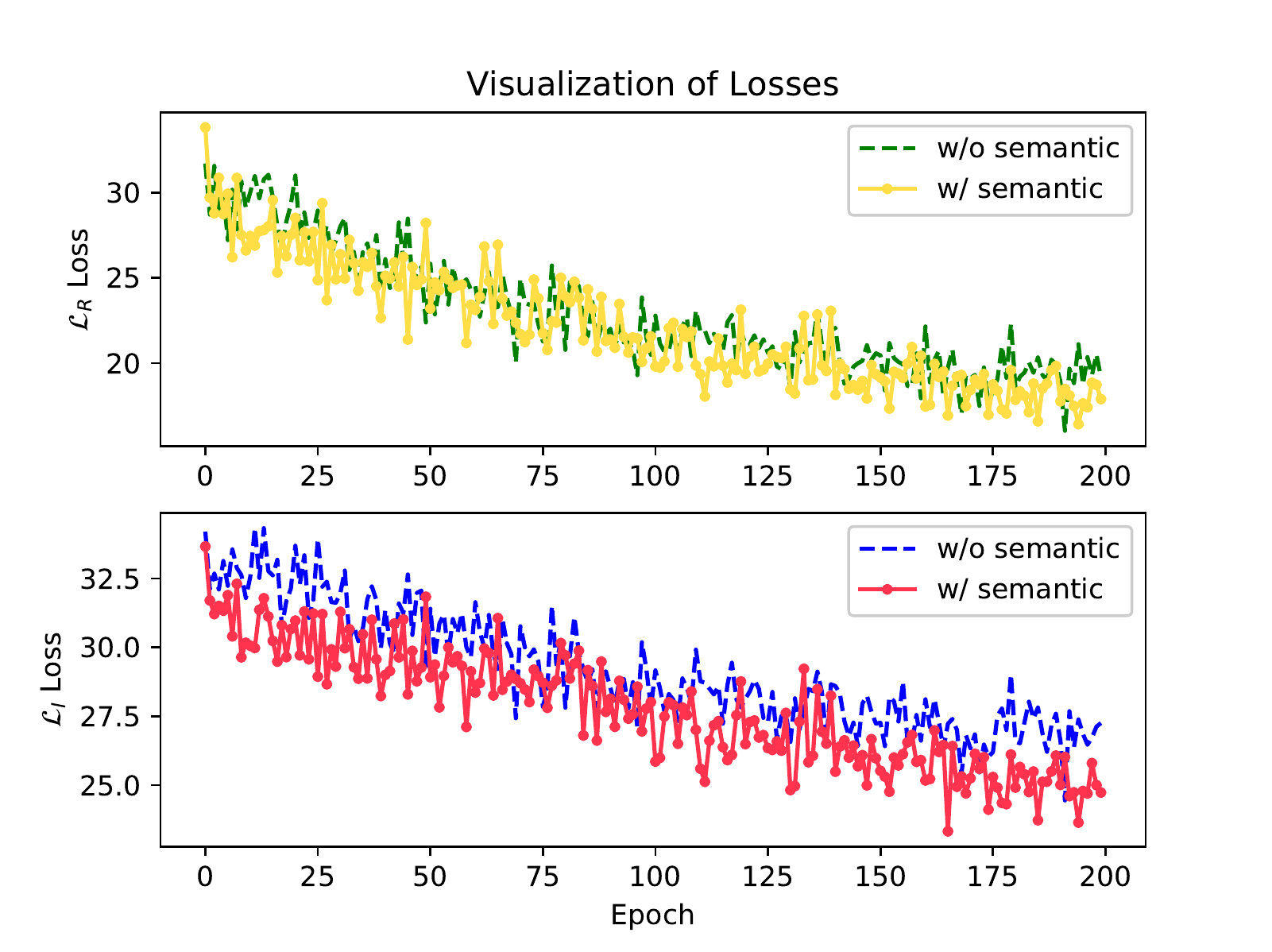}
        \caption{PCSO}
    \label{visualization_loss_pcso}    
    \end{subfigure}%
    ~ 
    \begin{subfigure}[t]{0.45\textwidth}
        \centering
        \includegraphics[width=0.9\textwidth]{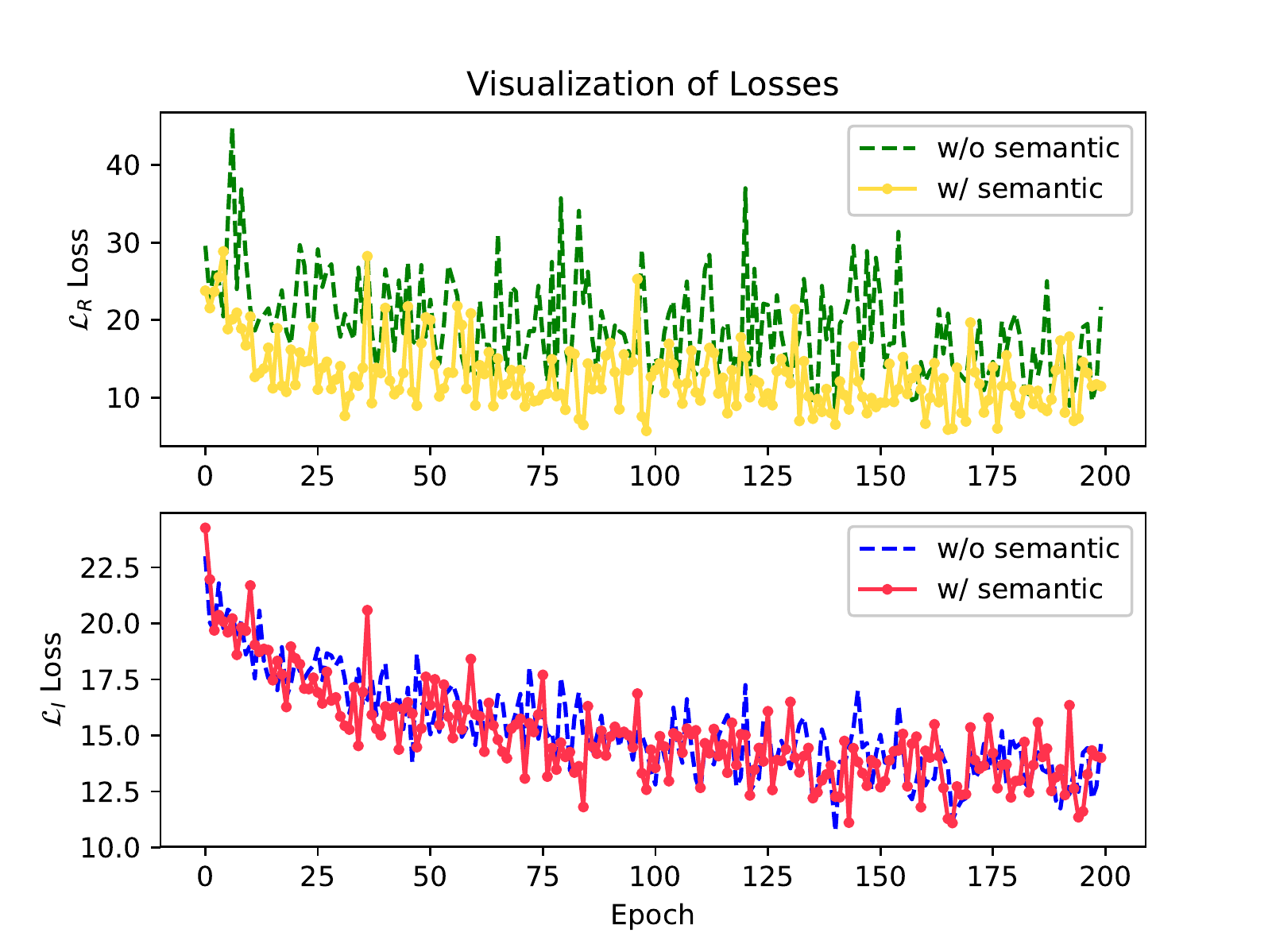}
        \caption{ARL}
     \label{visualization_loss_arl}
    \end{subfigure}
    \caption{Visualizing the convergence of different loss functions with and without the use of semantic regularization when training using the  PCSO and ARL datasets. The top row shows the $\mathcal{L}_R$ loss and the bottom row shows the $\mathcal{L}_I$ loss.}
    \label{visualization_loss}
    \end{figure*}
    
In addition to face matching experiments, we also computed and visualized the output images due to the use of $\mathcal{L}_R$ and $\mathcal{L}_I$ loss values on the same dataset. Loss values computed from individual batches were averaged during each training epoch. These two loss functions are vital to learning identity-specific features; hence,  their convergence reflects how well the GAN model is trained (see Figure~\ref{visualization_loss}(a)). 
%As seen from Figure~\ref{visualization_loss}(a), the use of semantic regularization ensures that the loss functions  $\mathcal{L}_R$ and $\mathcal{L}_I$ converge at a faster speed. 

The proposed method exhibits much better photo-realistic results in synthesizing THM from VIS face images.  This is of particular importance when cross-spectral face recognition systems deployed in practice require human intervention.  As can be seen from Figure~\ref{pcso_roc_matchers}, the effectiveness of the proposed SG-GAN method and the loss functions used are consistently observed across different face recognition matchers. This suggests that the proposed SG-GAN method is generalizable across different CNN-based face matchers.  For the evaluations below, the AM-Softmax matcher is adopted.

%-------------------------------------------------------------------------
\subsection{Evaluation on ARL Dataset}
\begin{figure}[h]
  \centering
    \includegraphics[width=0.45\textwidth]{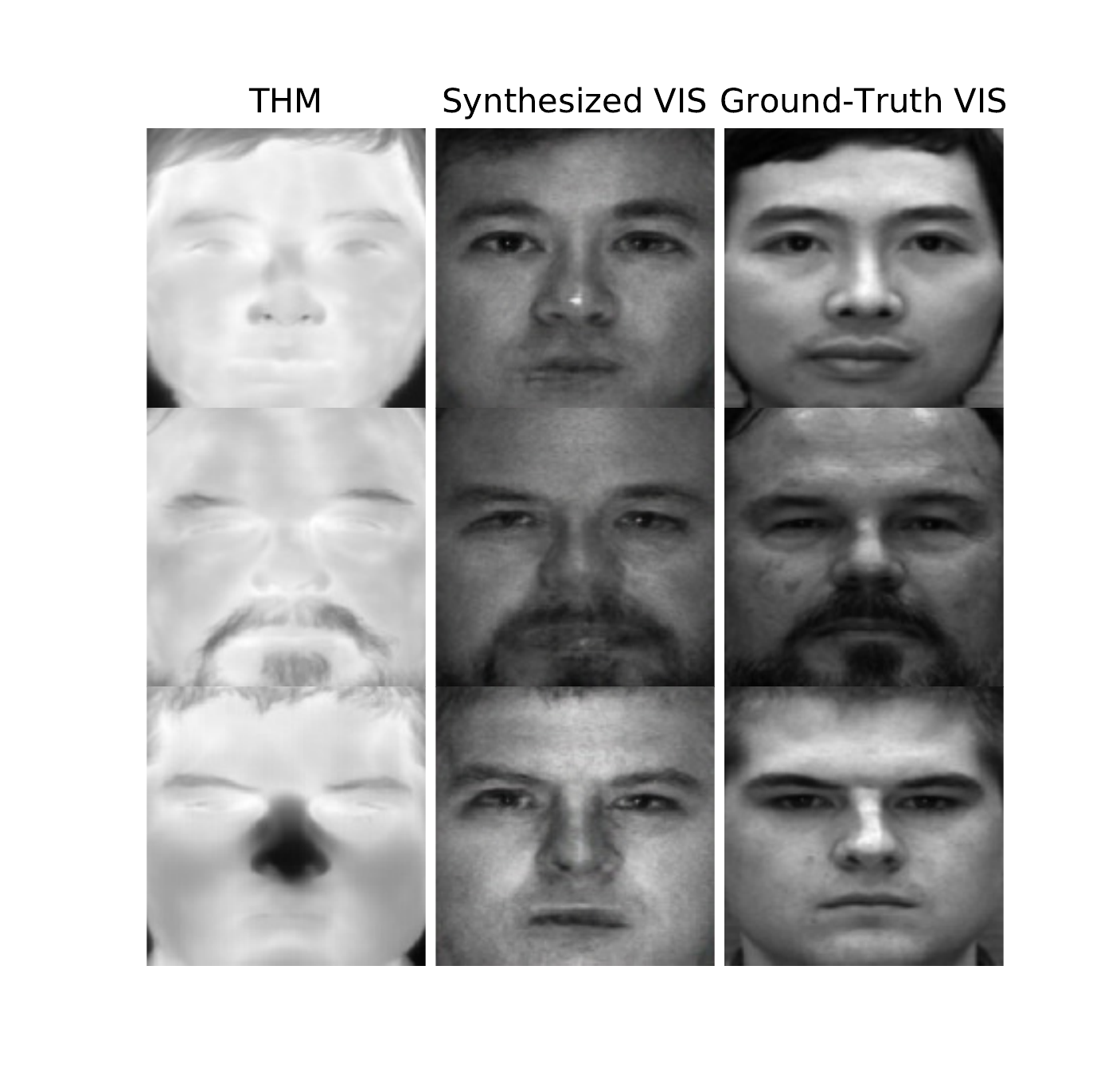}
    \caption{Generating VIS images from THM images on the ARL dataset using the SG-GAN method. }
    \label{arl_synthesized}
\end{figure}

The ARL dataset consists of thermal and visible face images captured from 60 subjects~\cite{HuDatabase2016}. According to the protocol used in~\cite{Riggan2016}, images of 60 subjects are used for the experiment. The interocular distance in these images is ~87 pixels. 30 subjects were randomly chosen for training and the remaining 30 were used for test and evaluation. This results in a total of 480 THM-VIS image pairs in the training and test sets. Face images in this dataset have a size of $250\times200$, which is center-cropped to a size of $200\times160$ by retaining a scale factor of of 0.8 in both vertical and horizontal directions, in order to remove unnecessary background information. Examples of images from the ARL dataset can be seen in Figure~\ref{arl_synthesized}. In this evaluation, we adopt 2-class semantic labels to compute the semantic loss function, resulting in better face matching performance.  This can be attributed to more robust semantic parsing formulation derived from the pre-trained face parsing network. We set $\lambda_S=20$ in this setting. 

\begin{figure}[h]
  \centering
    \includegraphics[width=0.45\textwidth]{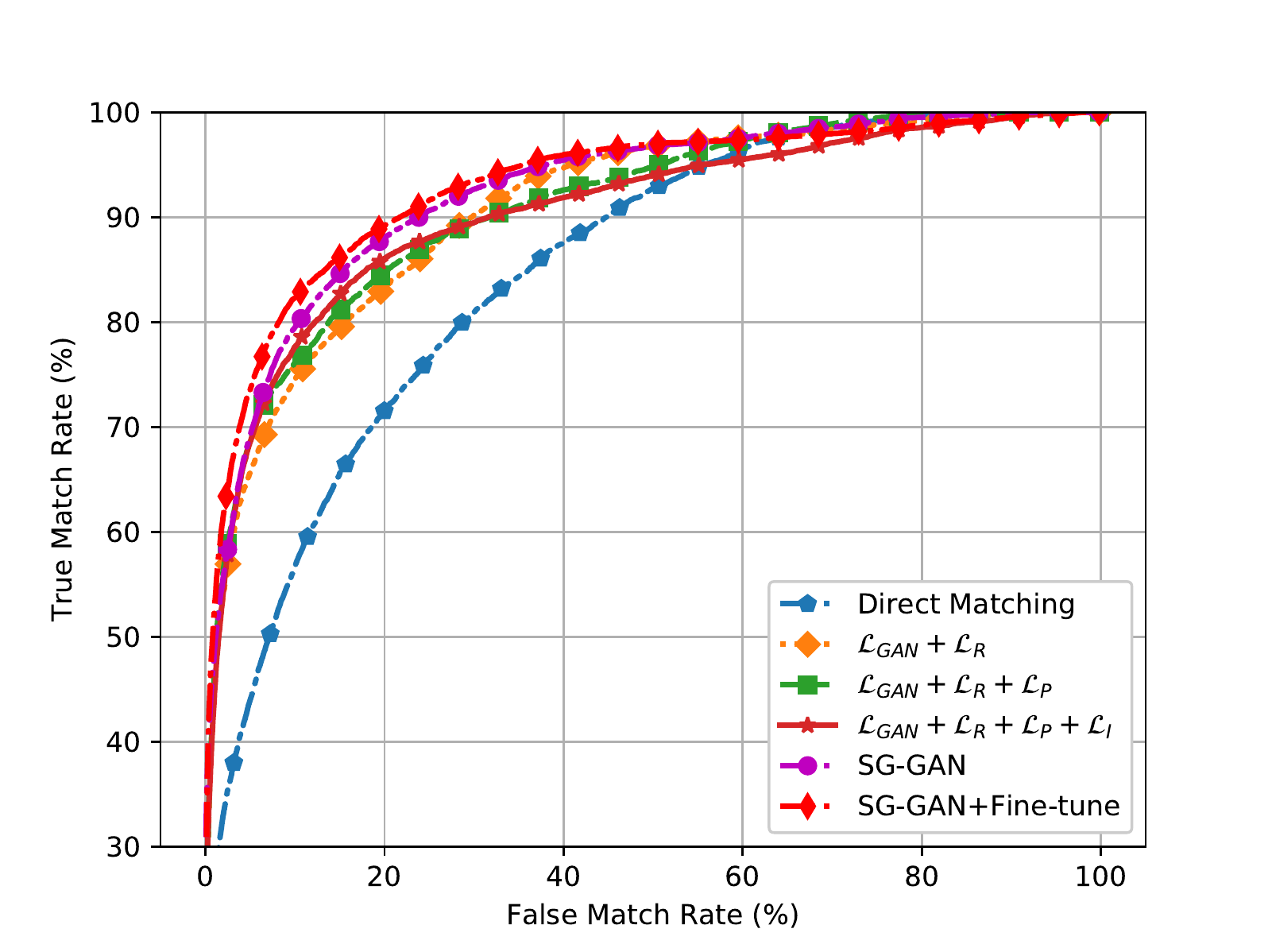}
    \caption{An ablation study of different loss functions on the ARL dataset. Our proposed solution  SG-GAN achieves much better performance than $\mathcal{L}_{GAN}+\mathcal{L}_R$. The face recognition matcher used here is AM-Softmax. The figure is best viewed in color.}
    \label{arl_roc}
\end{figure}

We also compare the proposed method against state-of-the-art CNN-based~\cite{Riggan2016, Riggan2018_region} and GAN-based~\cite{Zhang2017, Xing_2018_BTAS} synthesis methods that were previously evaluated on the ARL dataset. As seen from Table~\ref{ARL_results}, AP-GAN~\cite{Xing_2018_BTAS} using ground-truth attributes, rather than the estimated attributes, obtains 86.08\% AUC and 23.13\% EER, while our proposed SG-GAN method achieves 92.51\% AUC and 15.25\% EER. The AP-GAN~\cite{Xing_2018_BTAS} incorporates the attribute loss function when training its GAN; thus, the focus is on preserving attribute information. On the other hand, our proposed SG-GAN regularizes the training using semantic information; thus, the focus is on preserving facial information around significant components. An ablation study with respect to different combinations of loss functions is also conducted on this dataset (see Figure~\ref{arl_roc}). This further validates the effectiveness of our SG-GAN with semantic regularization. As seen from Figure~\ref{visualization_loss}(b), the use of such a regularization scheme ensures that the loss functions converge at a faster speed for $\mathcal{L}_R$. Though the convergence speed for $\mathcal{L}_I$ is not impacted, $\mathcal{L}_R$ is typically considered to be more significant for generating visually similar results since it operates at the pixel-level. 
\begin{table}[h]
\caption{Comparison of the proposed SG-GAN method against other state-of-the-art synthesis-based approaches previously reported on the ARL dataset. The AM-Softmax face matcher is used here. }
\begin{center}
  \begin{tabular}{ | c | c | c |}
    \hline
    \textbf{Algorithm} & \textbf{AUC (\%)} & \textbf{EER (\%)} \\ \hline
    Feature-based Synthesis~\cite{Riggan2016} & 68.52 & 34.36  \\ \hline
    Multi-Region Synthesis~\cite{Riggan2018_region} & 82.49 & 26.25 \\ \hline
    GAN-VFS~\cite{Zhang2017} & 79.30 &27.34\\ \hline
    AP-GAN~\cite{Xing_2018_BTAS} & 84.16 & 23.90 \\ \hline
    AP-GAN + GT~\cite{Xing_2018_BTAS} & 86.08 & 23.13 \\ \hline
    Proposed SG-GAN &92.51 &15.25\\ \hline
    SG-GAN + Fine-tune & \textbf{93.08} & \textbf{14.24} \\
    \hline
  \end{tabular}
\end{center}
\label{ARL_results}
\end{table}

The number of subjects in the ARL dataset is not as large as that of the PCSO dataset. This could potentially limit the capacity of the generator network to effectively learn features that are beneficial for thermal-to-visible  face synthesis. To address this issue, we utilize the SG-GAN model trained on the PCSO dataset and fine-tuned using the training partition of the ARL dataset. The trained SG-GAN model consists of two separate models: $G$ and $D$.  As described earlier, $G$ is responsible for generating photo-realistic VIS images and $D$ is used to determine whether the generated sample is real or synthesized. Two different fine-tuning strategies were tested. The first strategy performs fine-tuning on both $G$ and $D$ while the second strategy performs fine-tuning on $G$ only. Based on our experimental analysis, the latter is observed to result in better face matching accuracy. In the latter case, although we only use model $G$ for fine-tuning, model $D$ will continue to be trained based on the outputs from $G$.  As can be seen from Table~\ref{ARL_results} and Figure~\ref{arl_roc}, the fine-tuning approach obtains better results.  This offers a practical solution to deal with image synthesis using limited samples, since the fine-tuning strategy adopted in this work is different from the one used in image classification, where specific layers are fine-tuned. 

%------------------------------------------------------------------------
\section{Conclusions}\label{conclusion}
This paper proposes a novel synthesis-based method for matching thermal face images against visible spectrum images using a GAN-based approach. The proposed SG-GAN method utilizes semantic labels extracted by a face parsing network to compute the semantic loss function that regularizes network training, thereby improving both the quality of the synthesized face images and the accuracy of cross-spectral face matching. Experiments on two different datasets indicate that the proposed method is effective in synthesizing VIS images from THM images, and subsequently improves cross-spectral face matching accuracy. Future work would involve conducting experiments on a large-scale dataset to further investigate the role of different loss functions  in the proposed method.

%%%%%%%%%%%%%%%%%%%%%%%%%%%%%%%%%%%%%%%%%%%%%%%%%%%%%%%%%%%%%%%%%%%%%%%%%%%%%%%%
%\section{ACKNOWLEDGMENTS}

%The authors gratefully acknowledge the contribution of reviewers' comments, etc. (if desired). Put sponsor acknowledgments in the unnumbered footnote on the first page.

%%%%%%%%%%%%%%%%%%%%%%%%%%%%%%%%%%%%%%%%%%%%%%%%%%%%%%%%%%%%%%%%%%%%%%%%%%%%%%%%

%{\small
%\bibliographystyle{ieee}
%\bibliography{egbib}
%}

\end{document}